\documentclass[10pt,twocolumn,letterpaper]{article}

\usepackage{cvpr}
\usepackage{times}
\usepackage{epsfig}
\usepackage{graphicx}
\usepackage{amsmath}
\usepackage{amssymb}
\usepackage{subfigure,algorithmic,algorithm}
\usepackage[numbers]{natbib}
\usepackage{fancyhdr}

\newcommand{\figuredir}{./}
\newcommand{\figref}[1]{Fig.~\ref{fig:#1}}
\newcommand{\tblref}[1]{Table~\ref{tbl:#1}}

\newcommand{\secref}[1]{Sec.~\ref{sec:#1}}
\newcommand{\eqnref}[1]{Eq.~(\ref{eqn:#1})}
\newcommand{\q}[1]{\mathbf{#1}}
\newcommand{\qq}[1]{\mathbf{#1}}

\usepackage[pagebackref=true,breaklinks=true,letterpaper=true,colorlinks,bookmarks=false]{hyperref}

\cvprfinalcopy % *** Uncomment this line for the final submission

\def\cvprPaperID{****} % *** Enter the CVPR Paper ID here

\setcitestyle{authoryear}
\begin{document}

%%%%%%%%% TITLE
\title{Domain Adaptations for Computer Vision Applications}
\author{Oscar Beijbom, \\
{\tt\small obeijbom@ucsd.edu} \\
Department of Computer Science and Engineering, University of California, San Diego.}

\maketitle
\thispagestyle{empty}

%%%%%%%%% ABSTRACT
\begin{abstract}
A basic assumption of statistical learning theory is that train and test data are drawn from the same underlying distribution. Unfortunately, this assumption doesn't hold in many applications. Instead, ample labeled data might exist in a particular `source' domain while inference is needed in another, `target' domain. Domain adaptation methods leverage  labeled data from both domains to improve classification on unseen data in the target domain. In this work we survey domain transfer learning methods for various application domains with focus on recent work in Computer Vision.
\end{abstract}

% \tableofcontents
\section{Introduction}
The shortage of labeled data is a fundamental problem in applied machine learning. While huge amounts of unlabeled data is constantly being generated and made available in many domains, the cost of acquiring data labels remains high. Even, worse, sometimes the situation makes it highly impractical or even impossible to acquire labelled data (e.g. when the underlying distribution is constantly changing).

Domain adaptation (sometimes referred to as domain transfer learning) approach this problem by leveraging labelled data in a related domain, hereafter referred to as `source' domain, when learning a classifier for unseen data in a `target' domain. The domains are assumed to be \emph{related}, but not identical (in which case it becomes a standard machine learning problem). 

This situation occur in many domains. A few examples are: event detection in across video corpora from different domains (e.g. different tv - stations), named entity recognition across different text corpora (e.g. sports text corpus and news corpus), object recognition in images acquired in different domains (webcam versus Amazon stock photos). 

Domain adaptation (DA) only recently started receiving significant attention~\citep{daume_frustrating, ciprian_adaptation, hal_domain}, in particular for computer vision applications \citep{saenko_adapting, kulis_what, raghuraman_domain,i-hong_robust,lixin_domain,bergamo_exploiting}, although related field, such as covariate shift~\citep{hidetoshi_improving} has a longer history. It is perhaps indicative of the field being so new, that the proposed methods are of such different characteristic. To the best of our knowledge, there is only one previous survey of domain adaptation~\citep{jing_literature}, which focused on learning theory and natural language processing applications. Also~\citep{S_Jialin_A} did a thorough survey on the related field of transfer learning.

\subsection{Related Fields}
As mentioned in the introduction, the shortage of labeled data is a fundamental problem for applied machine learning. It is important enough that several areas of research is devoted to various aspects of this problem. In the active learning paradigm, labels are acquired in an interactive fashion to maximize the benefit of each new label~\citep{4408844}. Related approaches include~\citep{419}, where a `human-in-the-loop' determines which labels to update, thus making the `most' out of the acquired labels. Crowd sourcing through, e.g. Amazon mechanical turk (mTurk), allows for rapid collection of large amounts of labels, and much research is devoted to the efficient distribution of tasks and the interpretation and weighting of retrieved labels~\citep{398}. Further areas  include weakly supervised method, e.g. multiple instance learning~\citep{Dietterich199731} or latent structureal SVMs~\citep{Yu_learningstructural} where the level of supervision is lower than the given task demands. Other approaches include semi-supervised learning that make use of small amounts of labelled data together with large amounts of unlabeled data. Notably the concept of co-training~\citep{Blum:1998:CLU:279943.279962} is a popular approach.

More closely related to domain adaptation is transfer learning~\citep{S_Jialin_A}. In transfer learning (TL) the marginal distribution of the source and target data are similar but different tasks are considered. To make this problem tractable, it is typically assumed a common prior on the model parameters across tasks. A computer vision example is `one-shot learing'~\citep{1597116} where new visual categories are leaned using a single training example by leveraging data from other labelled categories. This is different from domain adaptation where the marginal data distributions of source and target are different, but the task is similar. 

Another related field is model-adaptation. Here, unlabeled data, sometimes referred to as \emph{background} data or \emph{auxiliary} data is used to regularize the class specific models. This paradigm has had much success in speaker verification~\citep{Douglas_A._Speaker}, and has also been applied to computer vision problems~\citep{dixit_adapted, L_Fei-Fei_Learning}. The methods used in this field, such as Adapted Gaussian Mixture Models~\citep{Douglas_A._Speaker} could trivially be used in a domain transfer setting by discarding source data labels, and letting the source data constitute the background model.

Cross-Modal classification / retrieval makes very similar assumptions on the data compared to DA but assume \emph{instance}, rather than \emph{class}, level relationship between the domains. In Cross-Modal classification, ample data from both domains are available at train time, and the unknown sample can come from any of the modalities. 

\section{Setup}
In this section we introduce notation and provide a overview of the paper.
\subsection{Notation}
\label{sec:notation}
Let X denote the input, and Y the output random variable. Let $P(X,Y)$ denote the joint probability distribution of X and Y. Let, similarly $P(X)$ and $P(Y)$ denote the marginal probability distributions. In the domain adaptation scenario, as mentioned in the introduction, we have two distinct distributions. Let $P_s(X,Y)$ denote the \emph{source} distribution where, typically, we have access to ample labelled data, and let $P_t(X,Y)$ be the \emph{target} distribution that we seek to estimate. We also let $P(X = x, Y = y) \equiv P(x, y)$ refer to the joint probability, thus differentiating it from P(X, Y) that represents the probability distribution.

Data is available from three sets: labelled data from the source domain 
$\mathcal{S}_l = \{(\mathbf{x}^s_i, y^s_i)\}_{i = 1}^{N^s}, \mathbf{x}^s \in \mathbb{R}^{d_s}$, drawn from a joint source probability distribution, $P_s(X,Y)$; labelled data from the target domain,
$\mathcal{T}_l = \{(\mathbf{x}^{t, l}_i, y^t_i)\}_{i = 1}^{N^{t,l}}, \mathbf{x}^{t,l} \in \mathbb{R}^{d_t}$, drawn from a joint target distribution, $P_t(X,Y)$; and unlabeled data,
$\mathcal{T}_u = \{(\mathbf{x}^{t, i}_i)\}_{i = 1}^{N^{t,u}}, \mathbf{x}^{t,u} \in \mathbb{R}^{d_t}$ from a marginal target distribution $P_t(X)$. It is commonly assumed $N^s \gg N^{t,l}$, and $d_s = d_t$. The target and source labels are generally assumed to belong to the same space, e.g. for the k-class classification task, $y^s, y^t \in \mathbb{Z}^k$. We further let $D^i$ be the data matrix for domain $i$, with one data sample per column, $i \in \{t,s\}$.

The goal of domain adaptation (DA) can thus be summarized as that of learning a function $\hat{y} = f(\mathbf{x}_u | \mathcal{D})$ that predicts the class, $y_u$ of an unseen sample from the target with high probability, $P_t(Y = \hat{y} | X = \mathbf{x}_u)$. $\mathcal{D}$ is different depending on the data assumptions. In supervised DA, $\mathcal{D} = \mathcal{S}_l \cup \mathcal{T}_l$, in unsupervised DA  $\mathcal{D} = \mathcal{S}_l \cup \mathcal{T}_u$, and semi supervised DA: $ \mathcal{D} = \mathcal{S}_l \cup \mathcal{T}_l \cup \mathcal{T}_u$.

\subsection{Overview}
As mentioned in the introduction domain adaptation is a relatively new field. It is also relatively loosely defined with regards to e.g. how `related' are the domains, and how `few' labelled samples exist in the target domain. Also, the general problem statement applies to several application domains, such as natural language processing and computer vision. For all these reasons, there is a big variety in the proposed methods. Inspired by the categorization proposed in~\citep{jing_literature}, we begin by considering instance weighting methods for relaxation of the DA assumptions in \secref{instance}. We consider methods utilizing the source data to regularize target models in \secref{priors}. We then survey method seeking common representation across domains in \secref{common}. Section \ref{sec:transfer} make connections to transfer learning, and \secref{mm} briefly survey method for multi-modal learning.

\section{Instance Weighting}
\label{sec:instance}
Following~\citep{jing_literature}, we first consider two relaxations of the DA problem. For the analysis we will use the empirical risk minimization framework proposed by ~\citep{vapnik_nature} for standard supervised data. Here we let $\theta \in \Theta$ be a model parameter from a given parameter space, and $\theta^*$ be the optimal parameter choice for the distribution $P(X,Y)$. Let further $l(x,y,\theta)$ be a loss function. In this framework we want to minimize
\begin{eqnarray}
\theta^* &=& \arg \min_{\theta \in \Theta} \sum_{(x,y) \in \mathcal{X} \times \mathcal{Y}} P(x,y)l(x,y,\theta)
\end{eqnarray}
$P(X, Y)$ is unknown but we can estimate it with the empirical distribution, $\tilde{P}(X, Y)$.
\begin{eqnarray}
 \tilde{\theta} &=& \arg \min_{\theta \in \Theta} \sum_{(x,y) \in \mathcal{X} \times \mathcal{Y}} \tilde{P}(x,y)l(x,y,\theta) \\
&=& \arg \min_{\theta \in \Theta}  \sum_{i=1}^Nl(x_i, y_i, \theta).
\end{eqnarray}
\citep{jing_literature} extend this to the DA problem and arrive at the following formulation
\begin{eqnarray}
\theta^*_t &=& \arg \min_{\theta \in \Theta} \sum_{(x,y) \in \mathcal{X} \times \mathcal{Y}} P_t(x,y)l(x,y,\theta) \\
 		&\approx& \arg \min_{\theta \in \Theta}  \sum_{i=1}^{N^s} \frac{P_t(x_i^t, y^t_i)}{P_s(x_i^s, y_i^s)} l(x^s_i, y^s_i, \theta). \label{eqn:da_full}
\end{eqnarray}
We see that weighing the loss of (source) training sample by $\frac{P_t(x, y)}{P_s(x, y)}$ provides a solution that is consistent with the empirical risk minimization framework. Clearly, if we had a good estimate of $P_t(X, Y)$ we would already be done, so this doesn't really help us, but the formulation is useful for the discussion below. In the following we consider two relaxations of the DA problem formulation. Class imbalance: $P_t(X|Y = y) = P_s(X|Y = y)$ and covariate shift: $P_t(Y|X) = P_s(Y|X)$.

\subsection{Class Imbalance}
One way to relax the DA problem formulation is to assume $P_t(X|Y) = P_s(X|Y)$, but $P_t(Y) \neq P_s(Y)$. This is called class imbalance, population drift or sampling bias. Consider, for example, training data sampled from a remote sensing application. Test data collected a at a later occasion may have different class distribution due to a changed landscape. 
Taking the assumptions into account, the ratio $\frac{P_t(x, y)}{P_s(x, y)}$ becomes
\begin{eqnarray}
\frac{P_t(x, y)}{P_s(x, y)} & = & \frac{P_t(y)}{P_s(y)}  \frac{P_t(x | y)}{P_s(x | y)} \\
& = & \frac{P_t(y)}{P_s(y)},
\end{eqnarray}
and we only need to consider $\frac{P_t(y)}{P_s(y)}$. This approach was explored in ~\citep{yi_support}. We can also re-sample the data to make the class distributions equal.

\subsection{Covariate Shift}
Covariance shift~\citep{hidetoshi_improving}, is another relaxation of DA. Here, given an observation, the class distributions are same in the source and target domains, but the marginal data distributions are different. $P_t(Y|X) = P_s(Y|X)$, but $P_t(X) \neq P_s(X)$. This situation arise, for example, in active learning, where the $P_s(X)$ tend to be biased to lie near the margin of the classifier. At a first glance, this situation appears not to present a problem, since $P_t(Y|X) = P_s(Y|X)$, which we can estimate from the data. Here is why it becomes a problem in practice. Assuming, first of all, that the model family we use is mismatched to the data, i.e. regardless of what parameter we choose the model won't fit the underlying distribution. Under this assumption, covariate shift becomes a problem for the following reason. The optimal fit of the source data will be such that it minimize model error in the dense area of $P_{s}(X)$ (because these areas will dominate the error). Now, since $P_{t}(X)$ is different from $P_{s}(X)$, the learned model will not be optimal for the target data (again, since the model family is mismatched). 

As in the previous section, $\frac{P_t(x, y)}{P_s(x, y)}$ can be simplified under these assumptions
\begin{eqnarray}
\frac{P_t(x, y)}{P_s(x, y)} & = & \frac{P_t(x)}{P_s(x)}  \frac{P_t(y | x)}{P_s(y | x)} \\
& = & \frac{P_t(x)}{P_s(x)}.
\end{eqnarray}
Again, a well founded solution can be identified by appropriate instance weighting of the loss function. ~\citep{hidetoshi_improving} explored this approach and show that the weighted model better estimate the data given a biased sampling function. The quantity $\frac{P_t(x)}{P_s(x)}$ can be estimated using \eg non-parametric kernel estimation~\citep{masashi_input, hidetoshi_improving}. ~\citep{Huang07correctingsample} proposed to directly estimate the ratio, i.e. the difference between the two distributions. They use the Kernel Mean Match,
\begin{align}
\Big | \Big | \frac{1}{N^s} \sum_{i=1}^{N^s} \beta_i \phi(\mathbf{x}_i^s) - \frac{1}{N^t} \sum_{i=1}^{N^t} \phi(\mathbf{x}_i^t) \Big | \Big |^2,
\end{align}
metric that measures the distribution distance in a Reproducing Kernel Hilbert Space.

\section{Source Distribution As Prior}
\label{sec:priors}
Often, the simplifying assumptions of the previous section doesn't hold. This section discuss method that use prior probabilities estimated on the source data to regularize the model. We first cover priors in the bayesian sense, and then some examples of discriminative methods.

% MEGA
\begin{figure}[t] 
\def \W {85mm}
\def \H {\W}
\centering
\includegraphics[width=\W]{\figuredir 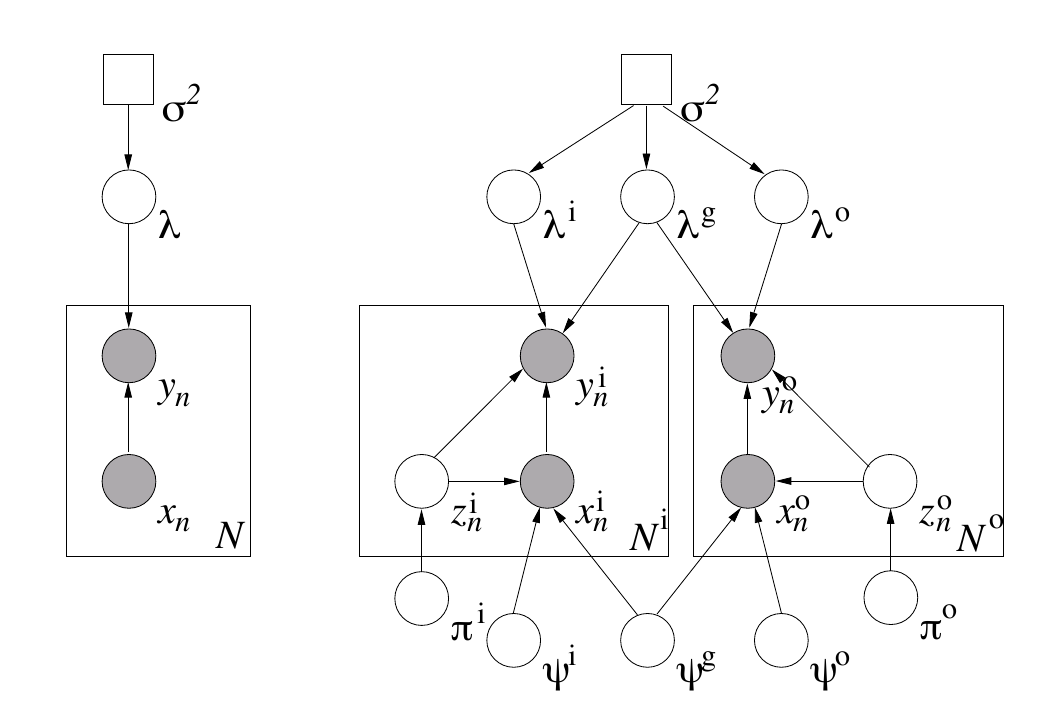}
\caption{The MEGA model proposed in \citep{hal_domain}. This model assumes the data is in fact generated by three distributions, a target, a common and and source. The MEGA model learns a classifier for each space. Left is the standard logistic regression model.}
\label{fig:mega}
\end{figure}
\subsection{Bayesian Priors}
\label{sec:bayprior}

Maximum a posterior (MAP) estimation of model parameters is central in bayesian statistics. In this setting prior knowledge about the model can be incorporated in a prior probability of the parameters, $P(\theta)$. Specifically, instead of finding optimal parameters $\theta^*$ as
\begin{eqnarray}
\theta^* = \arg \max_{\theta} \prod_{i = 1}^{n} P(y_i | x_i; \theta),
\end{eqnarray}
one solves
\begin{eqnarray}
\theta^* = \arg \max_{\theta} P(\theta) \prod_{i = 1}^{n} P(y_i | x_i, \theta).
\end{eqnarray}
In domain adaptation we can estimate the prior probability from the source domain as
\begin{eqnarray}
\theta^* = \arg \max_{\theta} P(\theta | \mathcal{S}_l) \prod_{i = 1}^{N^{t, l}} P(y^t_i | x^t_i, \theta).
\end{eqnarray}
\citep{ciprian_adaptation} pursued this approach in adapting a maximum entropy capitalizer. \citep{hal_domain} argued that this two step process (first estimating $P(\theta)$ from $S_l$ and then estimating $\theta$) was non-intuitive and suggested an ensemble model that considered three classifiers simultaneously, one for the target, one for the source and one for the joint portion of the data. This generative model, that they denote MEGA, is shown in \figref{mega}.

\begin{figure}[t] 
\def \W {85mm}
\def \H {\W}
\centering
\includegraphics[width=\W]{\figuredir 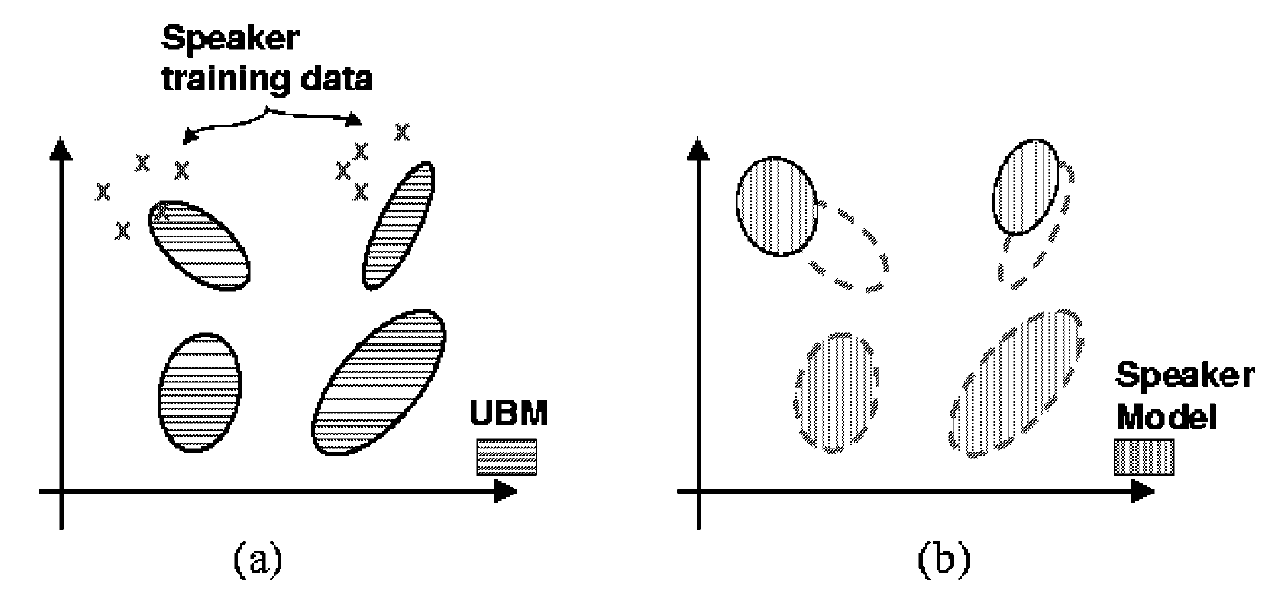}
\caption{Figure from~\citep{Douglas_A._Speaker} illustrating the adapted GMM model. The left figure shows the universal GMM estimated from the background data together with the speaker-specific train data. The right shows the adapted model.}
\label{fig:gmm}
\end{figure}

When $P(\theta | \mathcal{S})$ is estimated with unlabeled data, the problem is technically no longer domain adaptation but rather model adaptation. Adapted Gaussian Mixture Models have successfully been applied to speaker verification~\citep{Douglas_A._Speaker, W_M_Support}, and recently also for computer vision~\citep{dixit_adapted}. Figure \ref{fig:gmm} show a schematic illustration of adapted GMM.

% GMM

\subsection{Discriminative Priors}
In this section we survey work that investigate modifying the support vector machine (SVM) algorithms for the domain adaptation problem. These methods typically use a already-trained SVM in the source domain as input to subsequent training. The source data is thus used to \emph{regularize} the output model in a similar way as in \secref{bayprior}. 

\citep{jun_cross-domain} propose the adaptive support vector machine (ASVM). The basic idea is to learn a new decision boundary that is close to that learned in the source domain. The source data is thus acting as a regularizer on the final model. This method presume the existence of a SVM model $f^s(\mathbf{x})$ trained on the source domain data. They let the final decision function, $f(\mathbf{x})$ be the sum of $f^s(\mathbf{x})$ and $w^T \phi(\mathbf{x})$. The final classifier is attained by solving the following constrained optimization problem.
\begin{align*}
\min_{w} &\frac{1}{2} ||\mathbf{w}||_2^2 + C \sum_{i=1}^{N^{t, l}} \xi_i\\
s.t. 	~&\xi_i \geq 0 \\
	& y_i(f^s(\mathbf{x}_i) + \mathbf{w}^T\mathbf{x}^t_i) \geq 1 - \xi_i. 
\end{align*}
One problem with this formulation is that it doesn't strive for a large margin, but rather a solution close to the source solution. This is only reasonable for situation where $P_t(X,Y)$ is similar to $P_s(X,Y)$. To address this ~\citep{wei_cross-domain} proposed the Cross-Domain SVM (CDSVM). CDSVM relax the constraints that the final model need to be similar to the old one by only enforcing proximity where the support vectors of $f^s(x)$ are close to any of the target data. They do this by introducing additional constraints that the old support vectors, just like the target data points, should be correctly classified. These constraints are only active when the old support vectors are close to any part of the target data. Specifically, they solve the following constrained optimization problem
\begin{align}
\min_{w} & \frac{1}{2} ||\mathbf{w}||_2^2 + C \sum_{i=1}^{|\mathcal{T}_l|} \xi_i + C \sum_{j=1}^{M} \sigma(\mathbf{v}_j^s, \mathcal{T}_l) \bar{\xi}_j \notag \\
s.t.~& 	y_i(\mathbf{w}^T \phi(\mathbf{x}_i) - b) \geq 1 - \xi_i, \xi_i \geq 0, \forall(\mathbf{x}_i, y_i) \in \mathcal{T}_l \notag \\
	& y^s i(\mathbf{w}^T \phi(\mathbf{v}^s_j) - b) \geq 1 - \bar{\xi}, \bar{\xi} \geq 0, \forall(\mathbf{v}_j^s, y^s_j) \in \mathcal{V}_s. \notag 
\end{align}
Here $\mathbf{v}^s_j\in \mathcal{V}_s$ are the support vectors of $f^s(x)$ with signs $y^s_j$. The authors let $\sigma(\mathbf{v}^s, \mathcal{T}_l)$ be a gaussian that determines which vectors are close. \citep{bergamo_exploiting} provides a survey of other SVM-based DA methods.

\section{Common representation}
\label{sec:common}
The perhaps most intuitive way to do domain adaptation is to create a feature map such that the source and target distributions are aligned. In other words, finding functions $g_t(X)$ and $g_s(X)$ for which 
\begin{align}
P_t(Y=k | g_t(X) = x) = P_s(Y=k | g_s(X) = x) \notag \\
\forall (y, x) \in (\mathcal{Y} \times \mathcal{X})
\end{align}
where functions $g_t(X)$ and $g_s(X)$ might be equal, related or even identity, depending on the method.~\citep{jing_literature} note that the entropy of $Y | g(X)$ is likely to increase compared to $Y | X$ since the feature representation usually is simpler after the mapping, and thus encode less information. This means that Bayes error is likely to increase, and a good algorithm for domain alignment should take this into account.  
A simple, and straight forward way of doing this is by feature selection. ~\citep{Satpal_domainadaptation} proposed a method for this that remove features to minimize an approximated distance function between the source and target distributions. Specifically they minimize $\sum_{k \in K} d(E^k_{\mathcal{S}_l}, E^k_{\mathcal{T}_u})$ where 
\begin{align}
E^k_{\mathcal{S}_l} &= \sum_{(x_l, y_l) \in {\mathcal{S}_l}} \frac{f_k(\mathbf{x}_l, \mathbf{y}_l)}{N} \\
E^k_{\mathcal{T}_u} &= \sum_{(x_l) \in {\mathcal{T}_u}} \sum_y f_k(\mathbf{x}_l, \mathbf{y}) P(y|\mathbf{x}_l, \mathbf{w}), 
\end{align}
are the expectation of feature value $k$ for $\mathcal{S}_l$ and $\mathcal{T}_u$. The objective function is
\begin{align}
\arg \max_{\textbf{w}, K} \sum_{(x, y) \in \mathcal{S}_l} \sum_{k \in K} w_k f_k(\mathbf{x}, y) - \log z_w(\mathbf{x}) \\
s.t. \sum_{k \in K} d(E^k_{\mathcal{S}_l}, E^k_{\mathcal{T}_u}) \leq \epsilon
\end{align}

\citep{lixin_domain} took another aproach towards the same goal. They follow~\citep{borgwardt_integrating} and use the Maximum Mean Discrepancy (MMD) criterion, to compare data distributions based on the distance between the means of samples from the two domains in the Reproducing Kernel Hilbert Space (RKHS),
\begin{eqnarray}
\text{dist}_k(D^s, D^t) = \Big | \Big | \frac{1}{N^s} \sum_{i=1}^{N^s} \phi(\mathbf{x}_i^s) - \frac{1}{N^t} \sum_{i=1}^{N^t} \phi(\mathbf{x}_i^t) \Big | \Big |^2.
\end{eqnarray}
The authors integrate this distance with the standard SVM loss function
\begin{eqnarray}
[k^*, f^*] = \arg \min \Omega(\text{dist}_k(D^s, D^t)) + \sigma SVM_{k, f}(D),
\end{eqnarray}
thus jointly finding (1) a kernel that minimize $\text{dist}_k(D^s, D^t) $ and (2) a SVM decision function, $SVM_{k, f}$, that separate the data in kernel space. To make this tractable they iteratively solve for (the parameters of) a parameterized mixture of kernel functions and the ($\alpha$ parameters of) the SVM loss function. They show improvements on the TRECVID dataset over related approaches. 

\citep{John_Blitzer_Domain} proposed Structural Correspondence Learning (SCL). SCL finds a feature representation that maximize the correspondence between unlabeled data in source and target domain, by leveraging \emph{pivot} feature that behave similarly in both domains. For example, if the word on the right is `required' then the query word is likely a noun. This, then, helps disambiguate words such as 'signal', which can be both a noun and an adjective. This algorithm works on unsupervised data, and therefore doesn't maximize the correspondence between $P_t(Y | g_t(X))$ and $P_s(Y | g_s(X))$ directly, but rather on related tasks.

Several recent papers from the computer vision community pursue this idea. ~\citep{saenko_adapting}, and later~\citep{kulis_what} proposed variations on a metric learning formulation, where they not only learn a mapping that aligns the feature spaces but that also maximize class separation.
\begin{eqnarray}
\min_W r(W) + \lambda \sum_{i, j} c_{i, j}^W.
\end{eqnarray}
Here $D^t$ and $D^s$ are the target and source (labelled) data matrices respectively, with one sample per row. Saenko~\etal chose to $r(W)$ and $c_i()$ as,
\begin{align}
r(W) = &~\text{tr}(W) - \log \det(W) \notag \\
c_{i, j}^W  = &||x^t_i, x^s_j||_W \leq u ~|~y^t_i = y^s_j  \notag \\
 &||x^t_i, x^s_j||_W \geq l ~|~y^t_i \neq y^s_j, \label{kate1}
\end{align}
where $||a, b||_Q$ is the Mahalanobis distance between $a$ and $b$ with respect to matrix $Q$. With these constraints this formulations is known as information theoretic metric learning (ITML)~\citep{Davis07information-theoreticmetric}, and the algorithmic contribution of the paper is to enforce that each pair of datapoints are from the source and target domain, respectively. They state that this is crucial to ensure a domain transfer transform is learned.

 The authors note that since $\log \det(W)$ is only defined for positive definite matrices, one can decompose $W$ as $W = L^TL$. The mapping, therefore, is symmetric since $(D^t)^TWD^s = (D^t)^TL^TLD^s = (LD^t)^T (LD^s)$.~\citep{kulis_what}  address this by changing regularizer to the squared frobenius norm. They also changed the constraints to encode similarity of data samples rather than the Mahalanobis distance. The new formulation becomes
\begin{align}
r(W) = 		&~\frac{1}{2} ||W||^2_F \notag\\
c_{i, j}^W = 	&~\max(0, x^t_iWx^s_j - u) ^2 ~|~y^t_i = y^s_j \notag \\
			&~\max(0, l - x^t_iWx^s_j) ^ 2 ~|~y^t_i \neq y^s_j.
\end{align}
Kulis~\etal show how to kernelize this formulation. Their method show minor improvements on the `Saenko Items' dataset (\tblref{datasets}) compared to ~\citep{saenko_adapting, daume_frustrating} and baseline methods. 

~\citep{i-hong_robust} very recently proposed a formulation where the goal is to map the source data, by a matrix $W \in \mathbb{R}^{d \times d}$, to an intermediate representation where each transformed sample can be reconstructed by a linear combination of the target data samples,
\begin{eqnarray}
WD^s = D^tZ,
\end{eqnarray}
where $Z \in \mathbb{R}^{n^{t, l} \times N^s}$. They propose the following formulation to solve for low rank solutions.
\begin{align}
\min_{W, Z, E} &~\text{rank}(Z) + \alpha||E||_{2,1}, \notag \\
s.t. &~ WS = TZ + W, \notag \\
&~ WW^T = I.
\end{align}
 To solve this problem they relax the rank constraint to the nuclear norm and then apply a version of the Augmented Lagrange Multiplier (ALM) method~\citep{Lin_Chen_Ma_2010}.
 
Another recent method for computer vision also propose a mapping to a common representation~\citep{raghuraman_domain}. Motivate by incremental learning, they create intermediate representation between the source and domain data by viewing the generative subspaces created from these domains as points on a Grassmanian manifold. Intermediate representations can then be recovered by sampling the geodesic path. The final feature representation is a stacked feature vector, from each location along the path. They use partial least squares to learn a model on this extended feature representation. Table \ref{tbl:results} show the evaluation of this and several other discussed methods.
% MANIFOLD	
\begin{figure}[t] 
\def \W {85mm}
\def \H {\W}
\centering
\includegraphics[width=\W]{\figuredir 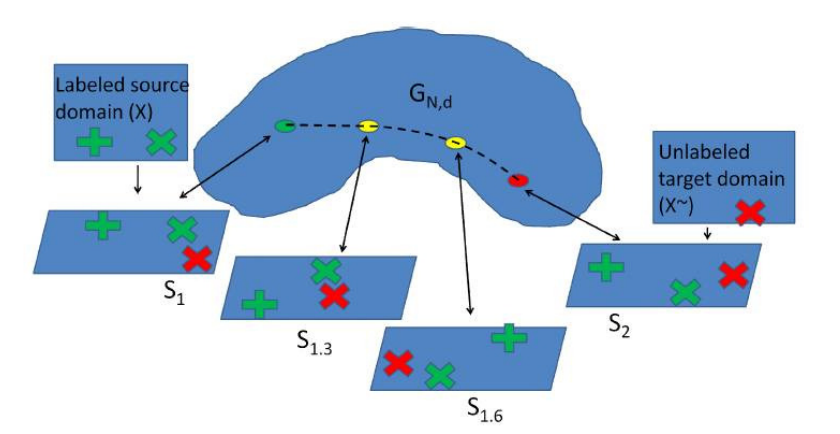}
\caption{Figure from~\citep{raghuraman_domain} illustrating the proposed method.}
\label{fig:manifold}
\end{figure}

%%% SAENKO OBJECT RESULT TABLE
\begin{table*} 
\begin{center}
\begin{tabular}{ | p{1.2cm} | p{1.2cm} | p{2cm} | p{1.2cm}| p{1.2cm} | p{2cm} | p{1.6cm} | p{2cm}| p{2cm}|}
	\hline 
	\multicolumn{2}{|c|}{Domain} & Naive - train on all data & \multicolumn{2}{|p{2.4cm}|}{Metric learning (Supervised) \citep{kulis_what}} & \multicolumn{2}{p{4cm}|}{Manifold \citep{raghuraman_domain}} & Low-rank (Supervised) \citep{i-hong_robust}  & A-SVM (Supervised) \citep{jun_cross-domain}\\
	\hline
	Source & Target & Naive & asymm & symm & Unsupervised & Supervised & RDALR & A-SVM\\
	\hline\hline
	webcam & dslr & 22.13 $\pm$ 1.2 &  25 & 27 & 19 $\pm$ 1.2 & \textbf{37 $\pm$ 2.3} & 32.89 $\pm$ 1.2 & 25.96 $\pm$ 0.7\\
	dslr & webcam & 32.17 $\pm$ 0.8 & 30 & 31 & 26 $\pm$ 0.8 & \textbf{36 $\pm$ 1.1} & \textbf{36.85 $\pm$ 1.3} & 33.01 $\pm$ 0.8\\
	amazon & webcam & 41.29 $\pm$ 1.3 & 48 & 44 & 39 $\pm$ 2.0 & \textbf{57 $\pm$ 3.5} & 50.71 $\pm$ 0.8 & 42.23 $\pm$ 0.9\\
 	\hline
\end{tabular}
\end{center}
\caption{Evaluation of discussed method on the DA dataset introduced in~\citep{kulis_what}. The naive method train on $\mathcal{S}_l \cup \mathcal{T}_l$. The methods are trained on 8 images per category (if source is \emph{webcam} or \emph{dslr} or 20 image (for \emph{amazon}) from the source domain and 3 per category for the target domain. The best result for each experiment is marked in bold.}
\label{tbl:results}
\end{table*}

\section{Transfer Learning}
\label{sec:transfer}
As mentioned in the introduction, transfer learning, sometimes called multi-task learning is different from DA. In transfer learning (TL) the joint probability of each task $\{P(Y_k, X)\}_{k = 1}^m$ are different but there is only one marginal data distribution $P(X)$. Normally, the state space of the $Y_k$ are assumed to be different, \eg $\Omega(Y_1) \neq \Omega(Y_2)$. When learning class conditional models, $\{P(Y_k | X, \theta_k)\}_{k = 1}^m$, it is typically assumed a common prior distribution of the variables $ \theta_1 \hdots \theta_k \sim P_{\Theta}(\theta)$.

DA, while formally different, can be thought of as a special case of transfer learning with two tasks, one on the source, and one on the target, where $\Omega(Y_s) = \Omega(Y_t)$.

The classic paper by~\citep{daume_frustrating} can be viewed in this framework. Daum{\'e} propose a simple feature space augmentation by
\begin{eqnarray}
\Phi^s(x) = \langle x,x,0 \rangle \\
\Phi^t(x) = \langle x,0,x \rangle
\end{eqnarray}
This `frustratingly easy' method show promising performance doing named-entity recognition on several text datasets. Under a linear classification algorithm, this is equivalent to decomposing the model parameters for class $k$ as $\sigma_c + \sigma_k$, where $\sigma_c$ is shared by all domains. This formulation is basically identical to the one proposed by~\citep{theodoros_regularized} for the purpose of transfer learning. The authors~\citep{daume_frustrating} provide a different analysis in the paper, where they argue the similarity to the method of ~\citep{ciprian_adaptation}. 

\section{Multi-Modal Learning}
\label{sec:mm}
In this section, we discuss the concept of multi-modal learning. In this setting, correspondences are assumed to be on instance, rather then category, level. Also, here it is commonly assumed ample train data is available in both domains. Similarly to \secref{common}, the common goal of most methods is to estimate transformations $L_t$ nad $L_s$ so that $P_s((X^T L_s = x | Y = k) = P_t(X^T L_t = x | Y = k)$. This can be done by letting \eg $L_s = I$, thus mapping the target domain to the source domain, or vice versa. One could also consider mapping both spaces into a common space. We will begin this section by reviewing Canonical Correlation Analysis, Principal Component Analysis, Linear Discriminant Analysis. We then consider recent work utilizing these methods~\citep{A_Sharma_Generalized}.

\subsection{Background}
In this section we recap the formulations of Principal Component Analysis (PCA), Linear Discriminant Analysis (LDA) and Canonical Correlation Analysis (CCA). 

\textbf{Principal Component Analysis:} PCA, is a popular dimensionality reduction method that projects the data into direction of maximum variance. It can be derived as follows. Let $x_1 \hdots x_n $ be the input data. Let $\q{w}_1$ be the desired projection direction. Let also $\q{w}_1^T\q{w}_1 = 1$. The mean of the projected data is $\q{w}_1^T\bar{x}$, where $\bar{x} = \frac{1}{N} \sum_i^N x_i$. The variance of the projected data is
\begin{equation}
var(x) = \frac{1}{N} \sum_i^N (\q{w}_1^Tx_i - \q{w}_1^T\bar{\q{x}})^2,
\end{equation}
which can be expressed in terms of the data covariance matrix,
\begin{equation}
S = \frac{1}{N} \sum_i^N(x_i - \bar{x})(x_i - \bar{x})^T,
\end{equation}
as $u^TSu$.
We now maximize the projected variance with respect to $w_1$. The constrained maximization problem can be written as
\begin{align}
\q{w_1} &= \arg \max_{\q{w_1}} \q{w_1}^T \Sigma_{tt} \q{w_1} \notag \\
& \mathrm{s.t.} ~\q{w_1}^T \q{w_1} = 1 \label{eqn:pca} 
\end{align}

\textbf{Linear Discriminant Analysis:} While PCA is very popular for unsupervised data dimensionality reduction purposes it is agnostic to class, and might project data in directions that are \emph{unsuitable} for class discrimination. Linear Discriminant Analysis (LDA) finds projection directions by minimizing the within class scatter matrix while maximizing the between class scatter. LDA can be derived as follows. Let $\mathcal{X}_1 = \{\q{x}_1^1, \hdots \q{x}^1_{l_1} \}$ and $\mathcal{X}_2 = \{ \q{x}_1^2, \hdots \q{x}^2_{l_2} \}$ be samples from two different classes. The data projection direction $\q{w_1}$ is given by solving
\begin{align}
\q{w_1} &= \arg \max_{\q{w_1}} \q{w_1}^T S_B \q{w_1} \notag \\
& \mathrm{s.t.} ~\q{w_1}^T S_W \q{w_1} = 1 \label{eqn:lda}
\end{align}
where 
\begin{align}
S_B := (\q{m}_1 - \q{m}_2) (\q{m}_1 - \q{m}_2)^T  \notag \\
S_W := \sum_{i = 1, 2} \sum_{\q{x} \in \mathcal{X_i}} (\q{x} - \q{m}_i) (\q{x} - \q{m}_i)^T 
\end{align}
are the within and between scatter matrices, and $m_i$ is the mean of samples in $\mathcal{X_i}$. Note the very similar forms of \eqnref{pca} and \eqnref{lda}. \eqnref{pca} is a regular eigenvalue problem, while \eqnref{lda} is a generalized eigenvalue problem.

\textbf{Canonical Correlation Analysis:} Developed by ~\citep{1936} CCA is a data analysis and dimensionality reduction method, that can be though of as a multi-modal extension to PCA. CCA finds basis vectors for two sets of variables such that the correlations between the projections of the variables onto these basis variables are mutually maximized. Using notation from \secref{notation}, we let $D^s$ be a $n^s$ by $d^s$ matrix with rows $x^s_i$, and $D^t$ similarly. CCA finds projection directions $w^s$ and $w^t$ to maximize the correlation between the projected data. More formally, it finds projection directions by solving the following optimization problem
\begin{eqnarray}
\rho & = & \max_{\qq{w^t}, \qq{w^s}} corr(D^t \qq{w^t}, D^s \qq{w^s})  \label{eqn:CCA}\\
& = & \max_{\qq{w^t}, \qq{w^s}} \frac{ \langle D^t \qq{w^t}, D^s \qq{w^s} \rangle} {||D^t \qq{w^t} || || D^s \qq{w^s}||} \\
 & = & \max_{\qq{w^t}, \qq{w^s}} \frac{\qq{w^t}^T \Sigma_{ts} \qq{w^s}} {\sqrt{w^T_x \Sigma_{tt} \qq{w^t} \qq{w^s}^T \Sigma_{ss} \qq{w^s}}}
\end{eqnarray}
where $\Sigma_{ij}$ is the covariance matrix between data in domains $i$ and $j$, $i,j \in \{s,t\}$. Note that this formulation requires the same number of samples from each domain, but not the same dimensionality. We also note that the optimization can be written as a constrained optimization as
\begin{eqnarray}
\max_{\qq{w^t}, \qq{w^s}} \left[ \begin{array}{c} \qq{w^t} \\ \qq{w^s} \end{array} \right]^T 
\left[ \begin{array}{c c} 0 & \Sigma_{ts} \\ \Sigma_{st} & 0 \end{array} \right] 
\left[ \begin{array}{c} \qq{w^t} \\ \qq{w^s} \end{array} \right] \label{eqn:CCAkan} \\
s.t. 
\left[ \begin{array}{c} \qq{w^t} \\ \qq{w^s} \end{array} \right]^T
\left[ \begin{array}{c c} \Sigma_{tt} & 0 \\ 0 & \Sigma_{ss} \end{array} \right] 
\left[ \begin{array}{c} \qq{w^t} \\ \qq{w^s} \end{array} \right] = 1
\end{eqnarray}
The optimization problem \eqnref{CCA} (or \eqnref{CCAkan}) can be formulated as a generalized eigenvalue problem which can be solved as efficiently as regular eigenvalue problems~\citep{David_R_Canonical}. For more details~\citep{David_R_Canonical} provides an excellent analysis on CCA and the kernelized version KCCA.

%\begin{eqnarray}
%\left[ \begin{array}{c c} 0 & \Sigma_{ts} \\ \Sigma_{zx} & 0 \end{array} \right] \left[ \begin{array}{c} \qq{w^t} \\ \qq{w^s} \end{array} \right] = 
%\lambda \left[ \begin{array}{c c} \Sigma_{tt} & 0 \\ 0 & \Sigma_{ss} \end{array} \right] \left[ \begin{array}{c} \qq{w^t} \\ \qq{w^s} \end{array} \right],
%\end{eqnarray}

\subsection{Generalized Multiview Analysis}
GMA was proposed by ~\citep{A_Sharma_Generalized} as a unifying framework for learning multi modal discriminative linear projections. They argue that methods such as PCA and LDA do not handle multi-view data. On, the other hand, methods such as CCA is not supervised. Other methods such as SVM-2K~\citep{Jason_DR_Two}, CDSVM~\citep{wei_cross-domain}, ASVM~\citep{jun_cross-domain} meet these criteria but do not generalize well to unseen classes. The unifying framework of GMA is

\begin{align} %%%% GMA %%%%
& \max_{\qq{w^t}, \qq{w^s}} \left[ \begin{array}{c} \qq{w^t} \\ \qq{w^s} \end{array} \right]^T 
\left[ \begin{array}{c c} A_t & \alpha Z_t Z_s^T \\ \alpha Z_s Z_t^T & \mu A_s \end{array} \right] 
\left[ \begin{array}{c} \qq{w^t} \\ \qq{w^s} \end{array} \right] \\
&s.t. 
 \left[ \begin{array}{c} \qq{w^t} \\ \qq{w^s} \end{array} \right]^T
\left[ \begin{array}{c c} B_t & 0 \\ 0 & \gamma B_s \end{array} \right] 
\left[ \begin{array}{c} \qq{w^t} \\ \qq{w^s} \end{array} \right] = 1.
\end{align}
PCA can be recovered (in the i'th view) through this framework by setting $A_i = \Sigma_{ii}, B_i = I$. Similarly, LDA can be recovered $A_i = S^B, B_i = S^W$, with $S^B$ and $S^W$ defined as above. CCA can be recovered as $A_i = 0, B_i = D^i W^i (X^i)^T$, and $Z_i = X_i$.

Using this framework they propose two methods, Generalized Multiview LDA and Generalized Multiview Marginal Fisher Analysis (GMMFA). Here we recap only GMLDA. As noted above, LDA in the i'th view can be achieved by setting $A_i = S^B, B_i = S^W$. By setting $Z_i = M_i$, a matrix with columns that are class means, they enforce class mean alignment across classes. The authors also note that the two step process of LDA + CCA or vice versa needs to be considered as a baseline method. Similar approaches include~\citep{rasiwasia_new} who introduced \emph{semantic correlation matching}, which uses logistic regression to combine CCA with semantic matching. They also introduce the WikiText data set (\tblref{datasets}). 

The proposed method outperforms all baseline methods on multiPIE and VOC2007 and is on par with the domain specific approach by ~\citep{rasiwasia_new} on the WikiText dataset.

%%% DATASET TABLE
\begin{table}[h] 
\begin{center}
\begin{tabular}{ | p{1.6cm} | p{3.5cm} | p{2cm} |}
	\hline 
	Name & Description & Instance or Class correspondence\\
	\hline\hline
	MultiPIE & Face recognition data set containing face images under different Pose Illumination and Expression & Instance \\
	 \hline
	WikiText & Each item is represented using a text and an image. & Instance \\
	 \hline
	Pascal VOC 2007 & 5011 / 4952 (training / testing) image-tag pairs & Instance \\
	 \hline
	Office dataset & Object images from Amazon, SLR and webcam & Class (subset with instance) \\
	 \hline
\end{tabular}
\end{center}
\caption{Computer vision datasets for domain adaptation method benchmark.}
\label{tbl:datasets}
\end{table}

{\small
\bibliographystyle{plainnat}
\bibliography{oscarsBibs.bib}
}

\end{document}